\documentclass[acmsmall]{acmart}
\usepackage{bm}
\AtBeginDocument{%
  \providecommand\BibTeX{{%
    \normalfont B\kern-0.5em{\scshape i\kern-0.25em b}\kern-0.8em\TeX}}}

\setcopyright{acmcopyright}
\acmJournal{TOMM}
\acmYear{2022} \acmVolume{1} \acmNumber{1} \acmArticle{1} \acmMonth{1} \acmPrice{15.00}\acmDOI{10.1145/3531017}

\usepackage{hyperref}

\newcommand{\hide}[1]{}

\usepackage{xcolor}
\usepackage{float}

\usepackage{multirow}
\usepackage{tcolorbox}

\newtcolorbox{mybox}[3][]
{
	colframe = #2!25,
	colback  = #2!10,
	coltitle = #2!20!black,  
	title    = #3,
	#1,
}

\usepackage{amsmath}

\usepackage{tabularx}
\usepackage{multirow}
\usepackage{subcaption}
\usepackage{enumitem}
\usepackage{bm}
\usepackage{soul}
\usepackage{arydshln}
\usepackage{algorithm}
\usepackage{algpseudocode}

\newcommand{\review}[1]{\textcolor{black}{#1}}

\begin{document}

\title{Interactive Garment Recommendation with User in the Loop}


\author{Federico Becattini}
\affiliation{%
  \institution{University of Siena}
  \country{Italy}}
\email{federico.becattini@unisi.it}

\author{Xiaolin Chen}
\email{e-mail: cxlicd@gmail.com}
\affiliation{%
  \institution{School of Software, Shandong University}
  \country{China}
  }

\author{Andrea Puccia}
\affiliation{%
  \institution{University of Florence}
  \country{Italy}}
\email{andrea.puccia@stud.unifi.it}

\author{Haokun Wen}
\affiliation{%
  \institution{School of Computer Science and Technology, Harbin Institute of Technology (Shenzhen)}
  \country{China}}
\email{whenhaokun@gmail.com}

\author{Xuemeng Song}
\affiliation{
  \institution{School of Computer Science and Technology, Shandong University}
  \country{China}}
\email{sxmustc@gmail.com}

\author{Liqiang Nie}
\affiliation{%
  \institution{School of Computer Science and Technology, Harbin Institute of Technology (Shenzhen)}
  \country{China}}
 \email{nieliqiang@gmail.com}
 
\author{Alberto Del Bimbo}
\affiliation{%
  \institution{University of Florence}
  \city{Florence}
  \country{Italy}}
\email{alberto.delbimbo@unifi.it}

\renewcommand{\shortauthors}{F. Becattini et al.}

\definecolor{cadmiumgreen}{rgb}{0.0, 0.42, 0.24}

\definecolor{linkblue}{rgb}{0.024,0.27,0.68}
\newcommand{\urllink}[1]{\color{linkblue}{\underline{\smash{\url{#1}}}}}

\begin{abstract}
Recommending fashion items often leverages rich user profiles and makes targeted suggestions based on past history and previous purchases. In this paper, we work under the assumption that no prior knowledge is given about a user. We propose to build a user profile on the fly by integrating user reactions as we recommend complementary items to compose an outfit. We present a reinforcement learning agent capable of suggesting appropriate garments and ingesting user feedback so to improve its recommendations and maximize user satisfaction. To train such a model, we resort to a proxy model to be able to simulate having user feedback in the training loop. We experiment on the IQON3000 fashion dataset and we find that a reinforcement learning-based agent becomes capable of improving its recommendations by taking into account personal preferences. Furthermore, such task demonstrated to be hard for non-reinforcement models, that cannot exploit exploration during training.
\end{abstract}





\maketitle

%%%%%%%%% BODY TEXT
\section{Introduction}
With the recent flourishing of the e-commerce fashion industry, the Internet has accumulated plenty of clothing data, making it increasingly difficult for users to seek their desired fashion items. Accordingly, fashion recommendation, which aims to recommend suitable items for users, has attracted considerable interest from both academia and industry due to its substantial economic value.
In fact, a surge of researches are dedicated to exploiting fashion recommendation~\cite{DBLP:conf/mm/GuanSZ0YC22,DBLP:conf/www/YuZHC0Q18}. 
However, in many cases, it is intractable for users to find the ideal items to exactly convey their intent. 
Therefore, to allow the users to flexibly provide feedback, interactive fashion recommendation merits our specific attention.

As a matter of fact, various methods~\cite{DBLP:journals/corr/abs-2207-12033,DBLP:conf/recsys/0001MO22, pal2023fashionntm} have been presented for interactive fashion recommendation, where the modification of the given visual image is regarded as user feedback provided as a textual description.
Despite the great success achieved by these efforts, they mainly work on \mbox{single-turn} modifications, and cannot further refine
the retrieved items in multi-turn.
Meanwhile, recent years have witnessed remarkable progress of refinement learning techniques in modeling dynamic evolution~\cite{DBLP:conf/cvpr/WuLZSZ22,DBLP:conf/aaai/YanaiSKSR22}, which propels us to explore the potential of incorporating refinement learning techniques in the context of interactive fashion recommendation.

\begin{figure}
    \centering
    \includegraphics[width=0.8\textwidth]{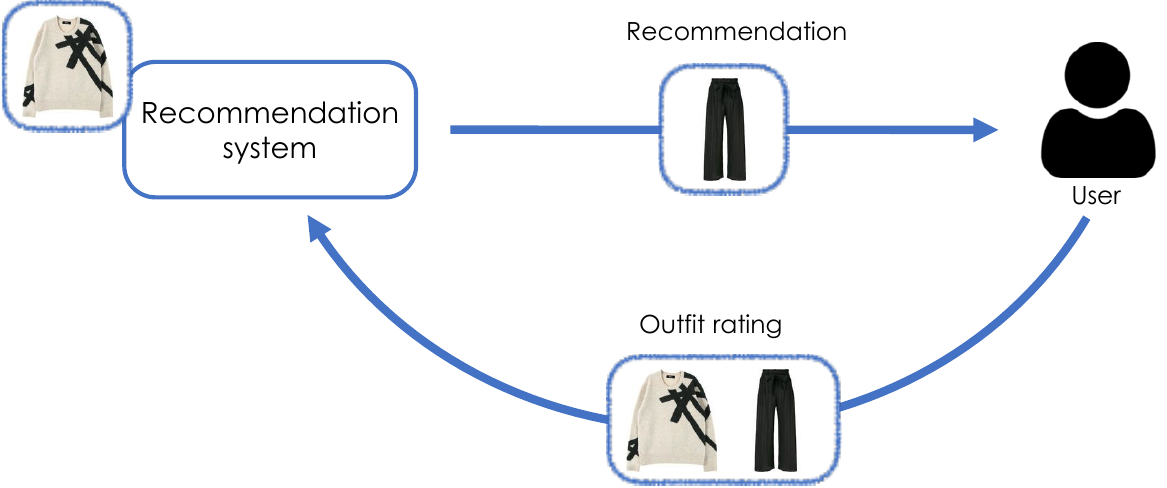}
    \caption{Compatible outfit recommendation through iterative feedback. The user presents a top garment to the system, which generates a recommendation, namely a bottom garment to be paired with the top. The user then provides feedback rating, according to personal preferences towards the generated outfit. The recommendation system integrates such feedback and proposes a new bottom to the user, attempting to maximize the feedback. The system does not have any prior knowledge about the user.}
    \label{fig:system}
\end{figure}

In this work, we aim to investigate the interactive fashion recommendation task, which allows users to express their feedback in a loop until the system recommends satisfactory fashion items. 
Fig. \ref{fig:system} shows an overview of the iterative recommendation system. When in operation, the system is initialized with a top garment given by the user. The system will then recommend bottom garments that are compatible with the top and such that the resulting outfit is satisfactory for the user.

Accurately fulfilling such interactive recommendation task is non-trivial. 
On the one hand, user feedback plays an essential role by providing personalized scores for different items and allowing continuous updating of the recommendation during the interaction. However, exploiting the user feedback for training is practically impossible due to the intractable size of data collections. According to this, some mechanism that subtly simulates the user behavior and provides a sufficiently accurate personalized score for each system proposal is needed.
On the other hand, the ultimate goal of providing such interactivity is making the system converge towards more and more satisfactory proposals. In this task, the system must keep track of all previously recommended items, by retaining an internal state that is updated after every iteration. The internal state has the responsibility to condense information in order to profile the user's feelings and provide appropriate recommendations at each timestep based on the current state. Training the system to make it able to capture the dynamic evolution of the dialogue and recommend the appropriate item is a tough challenge.

To address the aforementioned challenges, we propose a novel interactive system for fashion recommendation which consists of two key components: a \emph{User feedback model} and a \emph{Recommendation agent model}.
In more detail, for the first component, we exploit the GP-BPR~\cite{song2019gp} compatibility estimator as a proxy model of the user behavior. It estimates personalized outfit compatibility for each proposal and provides user feedback in the training loop.
For the second component, we employ reinforcement learning. A reinforcement learning agent performs garment recommendation in three steps: \emph{initialization}, \emph{recommendation}, and \emph{update}. Starting from the internal state, based on its learned policy, it determines the best action for the recommendation, \review{i.e., it selects the garment that is the most likely to be compatible with the input and the user}. We use $Q$-learning to predict the best action.
Extensive experiments have validated the effectiveness of our proposed system.

Our main contributions can be summarized as follows:
\begin{itemize}
    \item To the best of our knowledge, we are among the first to implement interactive fashion recommendation in a multi-turn manner, integrating feedback to maximize user satisfaction.
    \item We present a novel interactive fashion recommendation system, which employs \mbox{GP-BPR} as a proxy model to provide personalized scores for different items at training time. In addition, the proposed system utilizes a reinforcement learning agent to conduct the recommendation by keeping track of all previously suggested items.
    % allows users to express their feedback in a loop until the system recommends satisfactory fashion items. 
    \item Extensive experiments conducted on a real-world dataset demonstrate the effectiveness of our proposed system.
\end{itemize}

\label{sec:intro}

% ---------------------------------------------------------------------------------

\section{Related work}
\label{sec:related}
% Our work is closely related to the studies on fashion recommendation and reinforcement learning for recommendation. 
Our work is closely related to studies on fashion recommendation, feedback-guided fashion search, and reinforcement learning. 

\subsection{Fashion Recommendation}
Traditional recommendation systems utilizing collaborative filtering or content-based filtering are unsuitable in the fashion domain due to the sparsity of purchase records~\cite{DBLP:conf/waim/ShaWZFZY16}. Instead, the mainstream literature has leveraged the rich information of fashion items to achieve promising fashion recommendations based on fashion compatibility modeling, which estimates whether two or several fashion items go well together. For example, Song et al~\cite{DBLP:conf/mm/SongFLLNM17} proposed a multimodal compatibility modeling scheme, whereby the Bayesian Personalized Ranking optimization~\cite{DBLP:conf/csse/ZhaoW19} is adopted to model the compatibility between fashion items. Later, Vasileva et al.~\cite{DBLP:conf/eccv/VasilevaPDRKF18} considered the fashion item category information when studying the compatibility for the outfit. Additionally, Han et al.~\cite{DBLP:conf/mm/HanWJD17} regarded the outfit as a predefined ordered sequence of fashion items and employed the Bi-LSTM network to uncover outfit compatibility. Although this method can analyze the compatibility relationship of an outfit from the global perspective, it is questionable to represent an outfit as a sequence of ordered items. Considering this, the current mainstream methods model each outfit as a graph and turn to the graph neural networks~\cite{DBLP:conf/icml/NiepertAK16} to fulfill this task. For example, Cui et al.~\cite{DBLP:conf/www/CuiLWZW19} proposed node-wise graph neural networks towards fashion compatibility modeling. In this method, each outfit is represented as a graph, where each node denotes a specific fashion item. In addition, Guan et al.~\cite{DBLP:journals/tip/GuanWSWYCN22} developed a hierarchical outfit compatibility modeling based on \mbox{item-level} graph and \mbox{attribute-level} graph, which can perform compatibility modeling more comprehensively.
Similarly, in \cite{becattini2023transformer}, the authors combine the effectiveness of representing outfits using graphs and a transformer architecture, modifying its self-attention mechanisms by leveraging a graph-based message-passing mechanism.
Memory augmented neural networks have also been recently used for fashion recommendation. Such architectures can learn to store relevant items such as pairs of compatible garments \cite{de2023disentangling, de2021style, becattini2023fashion} or disentangled prototypes representing fashion attributes \cite{zhao2017memory, de2021garment, scaramuzzino2023attribute} and can also be used to manipulate features \cite{pal2023fashionntm}.

Although these studies have achieved significant success, they mainly focus on fashion compatibility modeling to recommend a compatible item to go well with the given items, while overlooking the users' individual demands and preferences. Beyond these studies, in this work, we explore the users' feedback information in the recommendation loop, so as to recommend the tailored fashion item.

\subsection{Feedback-guided Fashion Search}
Another line of work close to ours is feedback-guided fashion search~\cite{DBLP:conf/cvpr/Vo0S0L0H19}, where the users can utilize a modification text as feedback to refine the reference image to search for the ideal image. Since the most promising application scenarios for this task lie in the fashion domain, most existing efforts~\cite{val,clvcnet,cosmo,datir,dcnet,hetero,DBLP:conf/cvpr/BaldratiBUB22a,pal2023fashionntm} focus on tackling this task based on \mbox{fashion-oriented} datasets~\cite{DBLP:conf/cvpr/WuGGARGF21,DBLP:conf/nips/GuoWCRTF18,DBLP:conf/iccv/HanWHZZLZD17}. Many ingenious composition manners have been proposed to properly compose multimodal queries to capture users' search demands. For example, Vo et al.~\cite{DBLP:conf/cvpr/Vo0S0L0H19} first proposed a gating and residual model to compose the multimodal query features, where the reference image is adaptively preserved and transformed to meet the users' ideal image. Following that, Chen et al.~\cite{val} utilized the attention mechanism to fuse the multimodal query representations. Later, Wen et al.~\cite{clvcnet} utilized two \mbox{fine-grained} multimodal composition modules from both local and global perspectives, which can model the diverse modification demands more precisely. Recently, Baldrati et al.~\cite{DBLP:conf/cvpr/BaldratiBUB22a} resorted to adopting the pre-trained large model CLIP in this task, and brought out its potential to achieve \mbox{cutting-edge} performance.

Whereas, these efforts focus on the \mbox{single-turn} iterative fashion search, and cannot further refine the retrieved image in \mbox{multi-turn}. While in this work, we allow users to express their feedback in a loop until the system recommends satisfactory fashion items.

A few works have recently addressed multi-turn feedback, often including free text to precisely describe how recommendations should be refined, thus addressing the task as a sort of dialogue between the user and the recommender system \cite{yuan2021conversational, pal2023fashionntm}. Such systems are therefore required to include a natural language understanding framework, resulting in complex and slow architectures. Differently from these approaches, we do not attempt to follow some direct manipulation command or to implement language-based interactions but we rather strive to maximize the satisfaction of the user by modeling the provided feedback.

% \subsection{Reinforcement Learning for Recommendation}
\subsection{Reinforcement Learning}
As an active research topic, reinforcement learning works on training an agent to derive the optimal policy by interacting dynamically~\cite{DBLP:conf/cvpr/WuLZSZ22}, which has exhibited prominent success in a plethora of tasks, such as the person re-identification~~\cite{DBLP:conf/cvpr/WuLZSZ22} and basketball games modeling~\cite{DBLP:conf/aaai/YanaiSKSR22}. 
For example, Wu et al.~\cite{DBLP:conf/cvpr/WuLZSZ22} proposed a temporal complementarity-guided reinforcement learning approach for the image-to-video person re-identification task, which employs reinforcement learning to accumulate complementary information.
In addition, Chen et al.~\cite{DBLP:conf/aaai/YanaiSKSR22} presented a deep reinforcement learning based framework for modeling basketball games, where  reinforcement learning is employed to integrate
the discrete actions and continuous actions of basketball players.
In particular, various studies have also emerged in conducting recommendations based on reinforcement learning~\cite{DBLP:conf/www/ZhengZZXY0L18,DBLP:conf/sigir/XinKAJ20}. 
For example, Zheng et al.~\cite{DBLP:conf/www/ZhengZZXY0L18} conducted the online personalized news recommendation with a reinforcement learning framework, which can capture the dynamic nature of news characteristics and user preferences. 
Besides, Xin et al.~\cite{DBLP:conf/sigir/XinKAJ20} proposed self-supervised reinforcement learning for sequential recommendation tasks, where reinforcement learning is deployed to incorporate reward-driven properties.

In a sense, these studies have curated the superiority of reinforcement learning in modeling dynamic evolution, which inspires us to resort to reinforcement learning for modeling user engagement and thus fulfill the requirements of an iterative recommendation system.

%\todo{include a short summary of Q learning}

% ---------------------------------------------------------------------------------

\section{Problem Statement and Application Context}

% \todo{A few ideas, still in progress...}

\review{Here we outline our vision of an application context for an interactive recommendation framework.
Typically, a recommendation system to be effective must have a form of user engagement so that suggestions can be personalized.
In fact, a classic recommendation system uses data about a user's past behavior and preferences to make personalized suggestions. To gather this information, the system first builds a user profile based on data collected from the user's past interactions. This may include items that the user has viewed, liked, or purchased, as well as any ratings or feedback that the user has provided.
Once the system has built a user profile, it can use this information to make recommendations in a variety of scenarios. For example, if the user is actively looking for something, the system can leverage the user's profile to make targeted suggestions.
On the other hand, if the system wants to attract a passive user, it may use a call to action to encourage the user to engage with the recommendation system and discover personalized suggestions.
When the user is not known, i.e. when no user profile has been collected, the recommendations can still be refined based on how the user interacts in the current session.
In this case, user engagement is fundamental. This may involve prompting the user to provide ratings or feedback on the suggestions that the system makes or asking the user to specify their preferences or interests. By collecting active feedback from the user, the system can continuously improve the accuracy and relevance of its recommendations.
To summarize, a recommendation system can be deployed in two main scenarios: when a user is actively looking for something and when the system itself wants to attract a passive user.
In this paper we are interested in the latter.}

\review{In particular, we pose the question of how to attract customers with targeted recommendations when no information is known about the user. We imagine a scenario where unknown and uncooperative users may observe a smart shop window or advertising board and no explicit feedback can be gathered.
The objective is to make an effective call to action and stimulate the user to enter the shop or interact with the system itself, falling back to a classic recommendation system with an active user.
To reach this goal we assume to have access to some form of implicit feedback, which might not be actively provided by the user. Some preliminary studies on the matter have been recently carried out exploiting computer vision \cite{bigi2020automatic, becattini2021plm, becattini2022understanding, becattini2022mall}. In these approaches, a vision-based approach is used to estimate the degree of interest or appreciation of a user, based on their behaviors such as posture or facial (micro) expressions.
We imagine our proposed model to be framed in a scenario where vision-based modules are capable of effectively estimating human feedback without actively engaging them, but just by observing their body language. In this paper, however, we leverage a proxy model to estimate human reactions so to provide feedback to the recommendation model, without the need to have real humans in the loop, which would make training unfeasible.}

\begin{figure}
    \centering
    \includegraphics[width=0.9\textwidth]{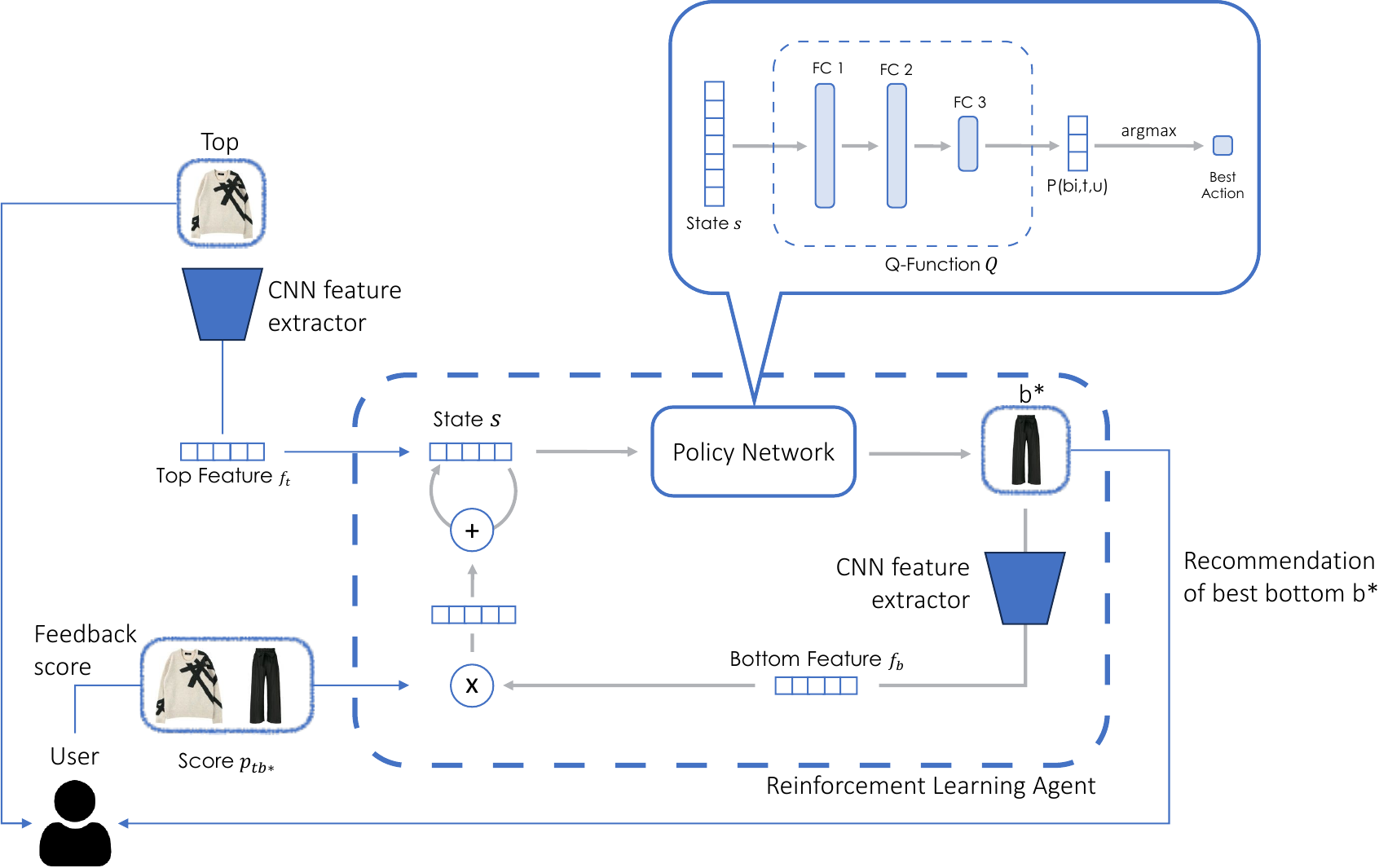}
    \caption{\review{Overview of the proposed approach. The system is initialized with the visual feature of the top garment presented by the user. The reinforcement learning agent, based on its policy, proposes an initial recommendation. The user then provides a feedback score, which is used to update the internal state of the agent. Based on such updated state, the agent proposes a new bottom, attempting to maximize the feedback score at every iteration. The Q-Function $Q$ selects the best action among the action space $A$. The function is approximated with a three-layer feedforward neural network, that projects the internal state of the agent into a distribution over all possible actions. The best action (i.e. the most compatible bottom) is selected by taking the most likely one according to the estimated likelihood.}}
    \label{fig:agent_model}
\end{figure}

\section{Interactive Garment Recommendation Model}
\label{sec:model_interative}
We propose a system that provides garment recommendations through a dialog with the user, where the user provides feedback for the proposed items and the system adjusts the next recommendation accordingly, with the intent of providing more satisfactory proposals.
We model feedback as a numeric evaluation, in order to maintain the system as generic as possible. Such score could be explicitly provided by the user through a digital signage display directly in the store, provided via a web interface in online shops, or could even be implicitly estimated from user reactions using computer vision techniques \cite{becattini2021plm, becattini2022understanding}.
By including the user feedback in the recommendation loop, the system is able to recommend personalized items without any prior knowledge of the user. As the model proposes a garment, the user evaluates the recommendation, thus the model is just optimized to maximize such feedback.
Fig. \ref{fig:agent_model} shows an overview of the iterative recommendation system.

The system operates in three steps as follows:
\begin{itemize}
    \item \textit{Initialization}: 
    let $t$ be a top garment and let's assume that the user is looking for a compatible bottom $b$. 
    The visual feature $f_t = F(t)$ of $t$ is extracted with a convolutional model $F$ and used to initialize the internal state of the reinforcement learning agent. Keeping track of the top garment in the agent's state allows the model to propose suitable bottoms.
    \item \textit{Recommendation}: 
    starting from its internal state, the agent selects the best action $a \in A$ according to some learned policy. \review{Each action corresponds to a different bottom garment being proposed to the user. The action space $A$ is composed of the set of possible recommendable bottom garments. Performing the best action therefore corresponds to generating a probability distribution over the set of candidate bottoms and selecting the garment with the highest probability.
    \begin{equation}
        a^* = argmax_i P(b_i | t, u)
    \end{equation}
    where $b_i$ is the i-th bottom in the collection, $t$ is the top and $u$ the user.}
    \item \textit{Updating}: \review{after a bottom $b*$ has been recommended, the model integrates the feedback from the user $p_{tb*}$ and updates its internal state $s$. The updating step takes the current internal state and modifies it by adding the bottom feature $f_b* = F(b*)$ of the proposed bottom, weighted by the feedback (\textit{compatibility score}) provided by the user. In this way, the system keeps track of all the recommended bottoms as a weighted sum of bottom features $f_b$:
    \begin{equation}
        s = s + p_{tb*} F(b^*)
    \end{equation}}
\end{itemize}
The \textit{Recommendation} and \textit{Updating} steps are repeated iteratively either for a finite number of steps or until the user is satisfied. To avoid repetitions in the \textit{Recommendation} step, candidate bottoms that have already been proposed are removed from the action space. \review{The procedure is detailed in Algorithm \ref{alg:algo}.}

\begin{algorithm}[t]
\caption{\review{Iterative recommendation algorithm. The system iteratively recommends garments to the user and integrates feedback scores on the fly.}}
\begin{algorithmic}[1]
    \State \textbf{Initialization}:
    \State \quad Let $t$ be a top garment.
    \State \quad Let $u$ be a user looking for a compatible bottom for $t$.
    \State \quad Visual feature extraction: $f_t = F(t)$, where $F$ is a convolutional neural network.
    \State \quad Agent internal state initialization: $s=f_t$.

    \State \textbf{Recommendation}:
    %\State \quad \textbf{for} $i$ \textbf{in} $1$ to $N$ \textbf{do}:
    \State \quad Generate probability distribution $P(b_i | t, u)$ for each candidate bottom $b_i$.
    %\State \quad \textbf{end for}
    \State \quad Select the action with the highest probability:
    \State \quad\quad\quad $a^* = \text{argmax}_i P(b_i | t, u)$
    \State \quad Recommend the corresponding bottom $b^*$
    
    \State \textbf{Update}:
    \State \quad Let $p_{tb*}$ be the feedback score given by user $u$ for the top $t$ and the recommended bottom $b*$.
    \State \quad \quad Integrate user feedback score and update internal state:
    \State \quad \quad $s \gets s + p_{tb*} F(b^*)$

    \State \textbf{Repeat} \textbf{Recommendation} and \textbf{Update} steps until max iter or user satisfaction.

    \State \textbf{Output}: Recommended bottoms based on user preferences.
\end{algorithmic}

\label{alg:algo}
\end{algorithm}

The recommendation system is implemented as a reinforcement learning agent. The agent has an internal state $s \in \mathcal{R}^N$, which has the same dimension as the features extracted from garment images with the convolutional model $F$. The agent can perform actions $a$, that are selected from an action space $A$. Any action corresponds to selecting a bottom garment from a set of possible candidates to be recommended to the user. 
The agent policy is learned by a Q-function that outputs the best action given the current internal state $s$. This is obtained according to the Bellman optimality equations for the Action-Value function \cite{sutton2018reinforcement}. In practice, the Q-function has been approximated by a neural network with three fully connected layers, as shown in Fig.~\ref{fig:agent_model}. The first two layers have 2048 neurons, while the third has a dimensionality equal to the cardinality of the action space, i.e. the number of possible bottoms. All intermediate layers have a ReLU activation. The network determines the best action by choosing the highest Action-Value function $Q(s, a)$ for a set of $n$ possible actions, given a state $s$.
As a reward, we use the difference between the user scores in two consecutive recommendations to enforce the model to cause positive feedback that are monotonically increasing through time.

% \begin{figure}
%     \centering
%     \includegraphics[width=0.8\textwidth]{img/qfunc.pdf}
%     \caption{Q-Function $Q$, that selects the best action among the action space $A$. The function is approximated with a three-layer feedforward neural network, that projects the internal state of the agent into a distribution over all possible actions. The best action is selected by taking the most likely according to the estimated likelihood.}
%     \label{fig:qfunc}
% \end{figure}

Including user feedback in the training loop is challenging. Ideally, for each user, we want to obtain a reward after each recommendation. While this is naturally obtained at inference time in real applications, it is unfeasible at training time, with thousands of samples and hundreds of users. Even assuming to have a dataset of outfits labeled with personalized scores, it remains impossible to ask each user to label any combination of top and bottom.
To overcome this limitation, we included a proxy model that replaces the user, providing compatibility scores for any top-bottom pair and any user in the training set. 
From the solutions proposed in literature \cite{DBLP:conf/mm/SongFLLNM17, DBLP:conf/csse/ZhaoW19, DBLP:conf/eccv/VasilevaPDRKF18, song2019gp} we have used  the state of the art GP-BPR model \cite{song2019gp} as the proxy-model of our system but, in principle, any personalized compatibility estimator could be used.

Using a proxy-model for training does not pose any limitations at test time. In fact, the recommendation system learns to optimize the recommendations based on the provided compatibility scores, without any knowledge of the source of such feedback.

\section{Training the system}
In the following, we discuss training for the different components of our system. However, before describing the training phase, we discuss the pre-processing steps that are needed to make processing effective.

\subsection{Preprocessing}
\begin{figure}[t]
	\centering
	\includegraphics[width=0.8\textwidth]{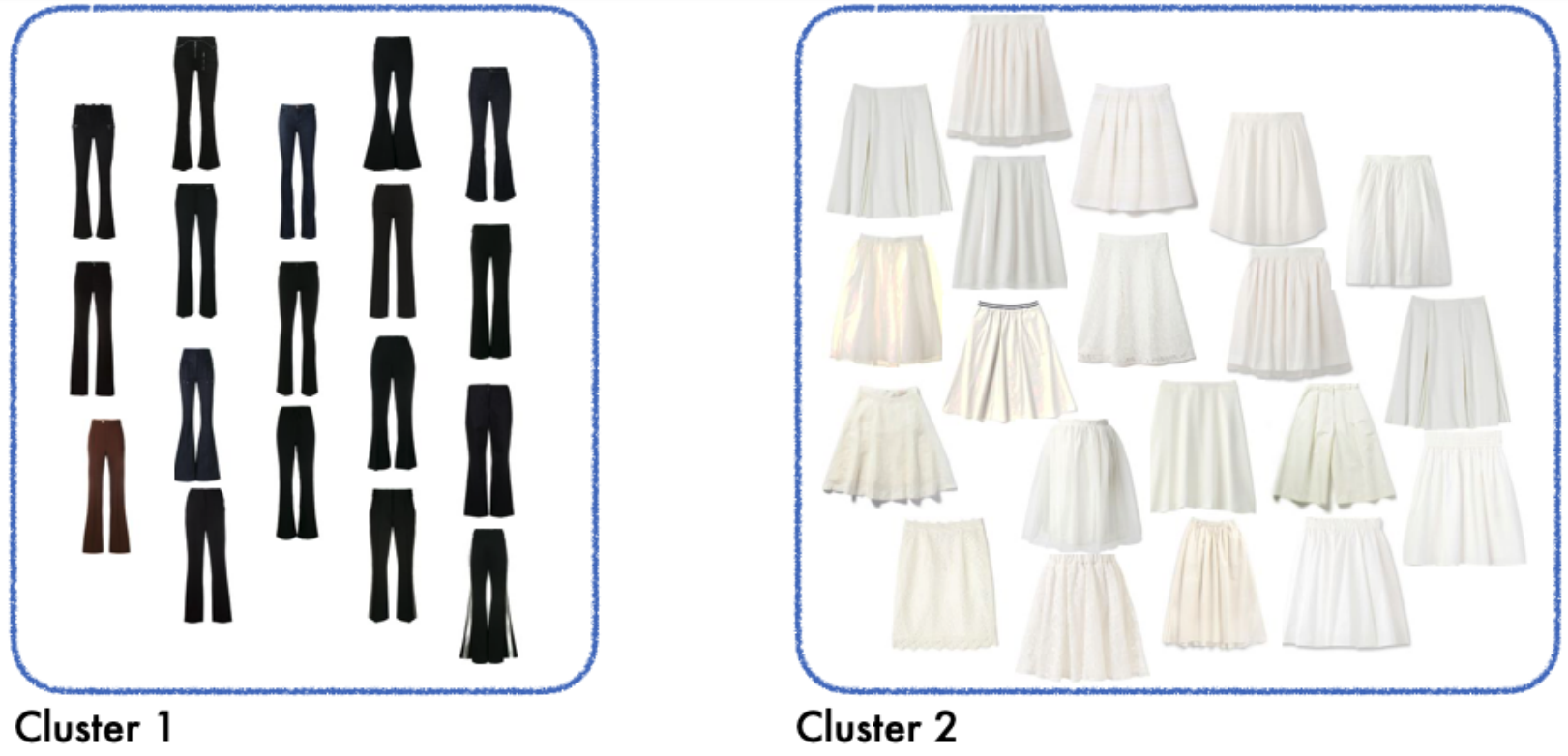}
	\caption{\review{We used K-means to group similar bottom garments before performing recommendations. These are two examples of clusters obtained from clustering images in the IQON3000 dataset. TO perform recommendation the model recommends a whole cluster by suggesting the garment that is closest to the centroid of the cluster.}}
	\label{fig:clusters}
\end{figure}

The rationale behind the need of pre-processing stems from the fact that the training dataset could include redundant items, namely variations of the same garments (e.g. blue jeans, white skirts, etc.). Since we strive to propose diverse recommendations to iteratively discover a garment that meets the needs of a user, we do not want to propose sequences of similar items. To this end, we group garments by similarity, and propose only one representative per cluster, allowing the user to fine-tune its search within the corresponding cluster afterwards. This also makes training easier, by reducing the complexity of the problem.

As a first step, we extract features using an Imagenet-pretrained ResNet model as in \cite{song2019gp} and then compress the features by training an autoencoder. This yields a compact visual representation of the items, on which to run a clustering algorithm.
The autoencoder takes as input the 2048-dimensional feature extracted by the last layer of ResNet and gradually projects it into smaller feature spaces of dimensions 1024, 512, 128 and 64 using fully connected layers with ReLU activations. The opposite is done to reconstruct the features using transposed convolutions.
We then run K-means on the 64-dimensional embeddings obtained with the autoencoder. \review{As the number of clusters we set $K=1000$}. An example of obtained clusters is shown in Fig. \ref{fig:clusters}.
In our framework we perform recommendations using cluster medoids, i.e. we consider only the closest element to the centroid for each cluster as recommendation candidates.

\subsection{User Proxy Model}
GP-BPR proposed by~\cite{song2019gp} aims to conduct the personalized compatibility modeling for personalized clothing matching, consisting of two pivotal components: \emph{general compatibility modeling} and \emph{personal preference modeling}. The general compatibility modeling component aims to calculate the general compatibility between a top garment and a bottom garment, while the personal preference modeling component devotes to modeling the user's preference towards the bottom garment for a given to-be-matched top garment.

{\textbf{General Compatibility Modeling.}} The general compatibility modeling component of GP-BPR aims to project the fashion garments into a latent space where the compatibility can be well reflected by distance metrics. 
%To be specific, this component first obtains the representation of fashion items. 
In particular, this component considers both the visual and contextual representations of each garment. \review{Notably, each garment is associated with an image and certain contextual information, like the brief description and category metadata.} Here, we take the visual representation leaning of the top garment as an example. Given the $i$-th top $v_i^t$, the representation can be calculated as follows,
% we can obtain its representation as follows,
 \begin{equation}
    \begin{split}
    \begin{cases}
    \mathbf{h}^t_{i1} = s(\mathbf{W}_1^t{\mathbf{v}_i^t}+\mathbf{b}_1^t),\\
    \mathbf{h}^t_{ik} = s(\mathbf{W}_k^t{\mathbf{h}^t_{i(k-1)}}+\mathbf{b}_k^t), k=2,\cdots,K,
    \end{cases}
    \end{split}
    \label{eq15}
\end{equation}
where $\mathbf{h}^t_{ik}$ refers to the hidden representation, $\mathbf{W}_k^t$ and $\mathbf{b}_k^t$, $k=1,\cdots,K$, are the weight matrices and biases. $s(\cdot)$ is the non-linear activation function in the multi-layer perceptron.
We treat the output of the $K$-th layer as the latent visual representation for the top garment~(\emph{i.e.,} $\mathbf{\tilde{v}}_i^t={\mathbf{h}^t_{iK}}\in \mathbb{R}^{D_{v0}}$), where ${D_{v0}}$ is the dimension of the latent space. Similarly, we can obtain the contextual representation for the top $t_i$, as well as the visual and contextual representations for the bottom $b_j$ as $\mathbf{\tilde{c}}_i^t$, $\mathbf{\tilde{v}}_j^b$, and $\mathbf{\tilde{c}}_j^b$, respectively. 
% Whereafter, we can calculate the general compatibility as follows,
Thereafter, the general compatibility score can be calculated as follows,
\begin{equation}
s_{ij} = \phi(\mathbf{\tilde{v}}_i^{t\top}){\mathbf{\tilde{v}}_j^b} + (1-\phi)({\mathbf{\tilde{c}}_i^t})^\top \mathbf{\tilde{v}}_j^b,
\label{eq:compatibility}
\end{equation}
where $\phi$ is the trade-off parameter, and $s_{ij}$ represents the general compatibility score between the top $t_i$ and bottom $b_j$.
\review{Note that the contextual information can be easily filtered out if textual meta-data is not available for the garments under consideration by simply setting $\phi$ to zero (i.e., discarding the second part of Eq. \ref{eq:compatibility}}

{\textbf{Personal Preference Modeling.}} The personal preference modeling component of GP-BPR resorts to matrix factorization to model user preference. To be specific,  the component firstly calculates the user preference $c_{mj}$ via the latent overall preference factors.
% as follows,
% \begin{equation}
% c_{mj} = \alpha + {\beta_m} +{\beta_j} + {\bm{\gamma}_m^\top}{\bm{\gamma}_j},
% \end{equation}
% where $c_{mj}$ is the preference of user $u_m$ for bottom $b_j$. $\alpha$ is the to-be-learned offset. $\beta_m$ and $\beta_j$ denote the user $u_m$ and bottom $b_j$, respectively.
Thereafter, the component also captures the latent content-based preference factors, which can be integrated into the matrix factorization
framework as follows,
\begin{equation}
c_{mj} = \alpha + {\beta_m} +{\beta_j} + {\bm{\gamma}_m^\top}{\bm{\gamma}_j} + \eta(\bm{\xi}_m^v)^\top{\bm{\xi}_j^v} + (1-\eta)(\bm{\xi}_m^c)^\top{\bm{\xi}_j^c},
\end{equation}
where $\bm{\gamma}_m$ and $\bm{\gamma}_j$ are the latent factors of user $u_m$
and bottom $b_j$, respectively.
$\bm{\xi}_m^v$ and $\bm{\xi}_j^v$ refer to the latent contextual factors of user $u_m$ and bottom $b_j$, while $\bm{\xi}_m^c$ and $\bm{\xi}_j^c$ denote the latent contextual factors of user $u_m$ and bottom $b_j$, respectively. $\eta$ is the non-negative
tradeoff parameter.

Ultimately, following the BPR loss, the GP-BPR framework has reached the final objective function for the personalized compatibility modeling as follows,
\begin{equation}
\mathcal{L}  = \sum_{(m,i,j,k)\in\mathcal{D}}{-ln(\sigma(p_{ij}^m-p_{ik}^m))} +\frac{\lambda}{2}||\bm{\Theta}_F||_F^2,
\end{equation}
% \begin{align}
% \mathcal{L} & = \sum_{(m,i,j,k)\in\mathcal{D}}{{l_{bpr}}(p_{ij}^m,p_{ik}^m)} +\frac{\lambda}{2}||\bm{\Theta}_F||_F^2,  \\& = \sum_{(m,i,j,k)\in\mathcal{D}}{-ln(\sigma(p_{ij}^m-p_{ik}^m))} +\frac{\lambda}{2}||\bm{\Theta}_F||_F^2,
% \end{align}
where $\mathcal{D}$ is the training set, and $(m,i,j,k)$ indicates that to match the given top $t_i$ and make a proper outfit, the user $u_m$ prefers the bottom $b_j$ to $b_k$.
$p_{ij}^m$ denotes the preference of the user $u_m$ towards the bottom $b_j$ for the top $t_i$, where $p_{ik}^m$ is the preference of the user $u_m$ towards the bottom $b_k$ for the top $t_i$.
$\lambda$ stands for the non-negative hyperparameter and the last term is designed to avoid the overfitting problem. $\bm{\Theta}_F$ refers to the model parameters.

\begin{figure}[t]
	\centering
	\includegraphics[width=0.6\textwidth]{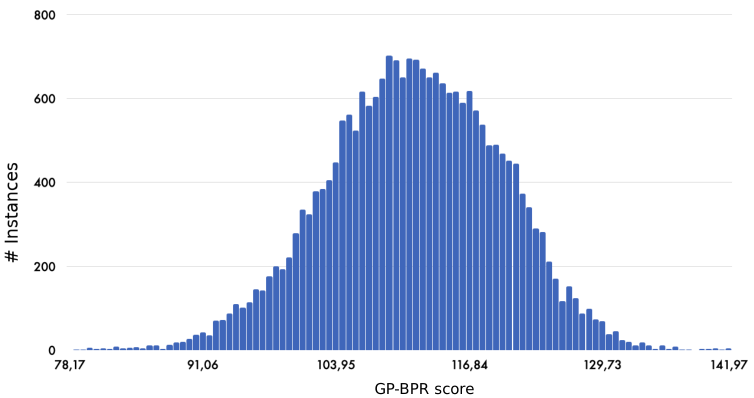}
	\caption{Unnormalized score distribution of GP-BPR on the validation set of IQON. We perform a min-max normalization to bound scores in [0, 1].}
	\label{fig:score_distribution}
\end{figure}

\textbf{GP-BPR Score Normalization.}
In this work, we rely on the GP-BPR model as a proxy to provide compatibility scores for user-outfit pairs.
The model provides unnormalized and unconstrained compatibility scores, yet since we use such a score in our feedback loop, we normalize it for better training stability. Furthermore, having normalized compatibility scores, e.g. in [0, 1] allows us to easily replace the proxy with a human user at inference time.

To normalize GP-BPR's scores, we compute the distribution of its compatibility scores on the validation set of the dataset we use for our experiments (IQON3000, see Sec. \ref{sec:dataset} for details). We show the distribution in Fig. \ref{fig:score_distribution}. We store statistics to perform min-max normalization at inference time. As min and max values, we used the 5th and 95th percentile of the distribution to filter outliers.

\subsection{Training the Interaction}

To train the network that approximates the Q-function, we rely on the Deep Q-Learning algorithm \cite{fan2020theoretical}, which is based on simulating the behavior of the agent.
The training process undergoes the following steps: (i) the simulation is initialized by selecting a top garment and a user at random; (ii) the agent acts following a training policy and, at every step, information is gathered regarding current state, chosen action, reward and next state; (iii) previously observed instances are sampled from a Replay Memory \cite{lin1992self} in order to compute the cost function necessary to apply a gradient descent step and update the neural network.

The role of the training policy is fundamental for training since it guarantees the right compromise between \textit{exploration} and \textit{exploitation}. The agent, in fact, must explore the state space and the action space to learn an effective Q-function for each state-action pair. At the same time, actions that allow the agent to accumulate rewards by simulating its real behavior must be chosen. For this reason we adopt an $\epsilon$-greedy training policy where, with probability $\epsilon$ a random action is chose (\textit{exploration}) while with probability $1-\epsilon$ the action that maximises the Q-function is chosen (\textit{exploitation}).

The value of $\epsilon$ balances exploration and exploitation. During training we used an exponential decaying $\epsilon$ in order to converge to a greedy policy while guaranteeing exploration during an initial training phase.
At each epoch $i$ we update $\epsilon$ in the following way:
\begin{equation}
\label{eq:epsilon}
\epsilon = \epsilon_{end} + (\epsilon_{start} -\epsilon_{end}) \cdot e^{\frac{-i}{\epsilon_{decay}}} 
\end{equation}
where $\epsilon_{start}$ and $\epsilon_{end}$ are respectively the initial and ending $\epsilon$ values and $\epsilon_{decay}$ regulates the decaying step.
\review{We use $\epsilon_{start}=0.9$, $\epsilon_{end}=0.25$ and $\epsilon_{decay}=200$}.

%\todo{We still have to try the RL model using the GAN features \url{https://github.com/FangShiting/iterative-recommendation/tree/master/GPBPR_GAN}}

% ---------------------------------------------------------------------------------

\section{Experimental setting}
\label{sec:exp}

% --------------------------------------

\begin{table}[!t]
\caption{Detailed statistics of the IQON3000 dataset.}
\begin{tabular}{|l|l|l|l|}
\hline
Category  & Number  & Category    & Number   \\ \hline
Outerwear & 35, 765 & Top         & 119, 895 \\ \hline
Bottom    & 77, 813 & Shoes       & 106, 598 \\ \hline
One Piece & 25, 816 & Accessories & 306, 448 \\ \hline
\end{tabular}
\label{dataset}
\end{table}

\subsection{Dataset and metrics}
\label{sec:dataset}
In this work, we adopted the IQON3000 dataset~\cite{DBLP:conf/mm/SongHLCXN19}, which contains $308,747$ outfits created by $3,568$ users with $672,335$ fashion items. To be specific, items in IQON3000 span six common categories: \emph{Coat, Top, Bottom, One Piece, Shoes and Accessories}, and as for each item, there is the corresponding visual image, categories, attributes, and description. Besides, each outfit is associated with its price and number of likes. For each outfit in the dataset, there are also examples of negative garments that would not pair well with the others in the outfit.
More detailed information about the IQON3000 dataset is summarized in Tab.~\ref{dataset}.
In this work, we only retain data regarding top and bottom pairs, leaving the development of a recommendation system for a whole outfit including different categories as future work. In particular, we focus on recommending bottom garments to be paired with a given top, although it should be noted that swapping tops and bottoms is trivial.

\review{The experimental setting is the following. We organize our experiments in episodes, corresponding to a recommendation session. At the beginning of each episode, we assume to have a user id and a top id, corresponding to the garment for which the user wants to find a compatible bottom. Our model can then perform a sequence of $N$ recommendation steps, incorporating user feedback at the end of every iteration. Ideally, the model should be able to maximize the feedback, i.e. the satisfaction of the user. In our experiments we use $N=10$.}

To evaluate our experiments, we resort to several metrics, as defined as follows.
To compare relative improvements in user feedback we leverage raw GP-BPR scores since they provide accurate enough estimates of how a given user (observed at training time) can respond to a top-bottom pairing. Ideally, one would want monotonically increasing scores, indicating that the model is suggesting increasingly better items.

We then introduce two metrics to quantify the quality of the recommendations, \textit{Higher-than-Negative} (HN) and \textit{Higher-than-Positive} (HP). These two metrics refer to the compatibility score (as provided by GP-BPR) obtained with the triplet composed of the given user, the given top and a candidate bottom. In particular, we define HP as 1 if at least one of the suggested bottoms obtains a score which is higher than the ground truth bottom, and zero otherwise. Vice-versa, we define HN as 1 if at least one candidate bottom is considered to be better than the negative bottom associated with the outfit, and zero otherwise.
In simple words, the HN metric indicates whether the suggested item is better than a negative, random bottom. This metric acts as a control experiment, indicating whether the system is suggesting random items or not. On the other hand, HP indicates whether we are able to suggest at least one item which is even better than the positive reference given by the ground truth. Note that the ground truth bottom does not necessarily achieve the highest score among all possible bottoms. This is because outfits in the dataset are composed of garments that are considered compatible by the user, without excluding the possibility that the user might prefer another item to improve the overall outfit.

We also report variants of these metrics, namely HN@T and HP@T, indicating if there is at least one recommended item that is better than the associated positive element (in case of HP) or negative element (in case of HN) within the first T recommendations.

We also report two additional metrics. First, we compute the overall number of proposed bottoms across the whole test set. This metric is important to assess if the model has the tendency to recommend always the same outfits and therefore has been affected by biases in the data. Ideally, the model should be able to recommend a large amount of items, depending on the input data and the feedback of the user.
Finally, we also report the overall amount of recommendations that obtain a score higher than the negative reference in the dataset.

% --------------------------------------

% --------------------------------------

%\subsection{Baselines}
%\todo{Do we have any simple baselines we can provide?}

% ---------------------------------------------------------------------------------

\section{Results}
\label{sec:res}
%\todo{Results}

\begin{figure}[t]
    \centering
    \includegraphics[width=0.65\textwidth]{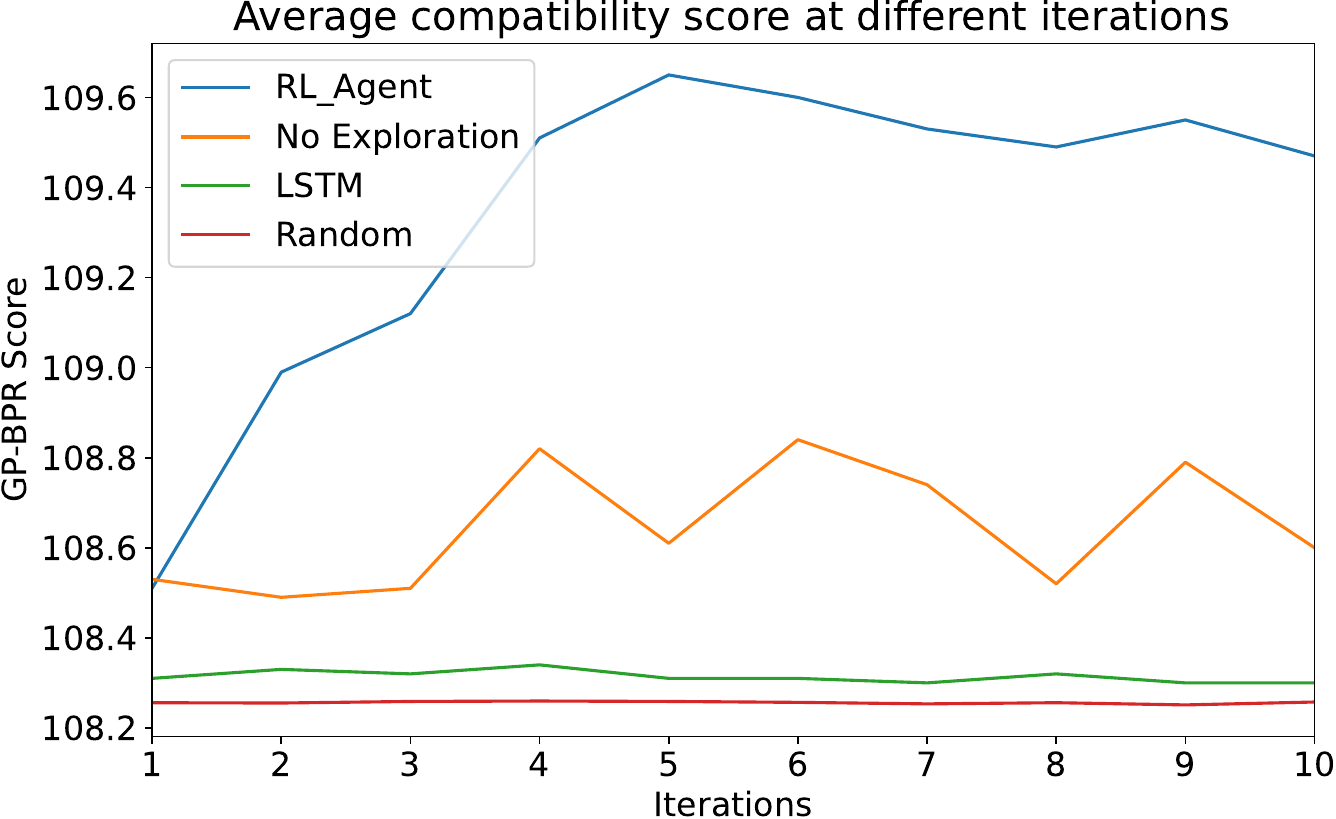}
    \caption{Average GP-BPR scores obtained by the recommendation system at different iterations. At first, the system suggests a garment without knowing any prior information about the user. Therefore, such a recommendation cannot take into account personal preferences. As the user starts providing feedback, the model is able to suggest on average a better item to complement the given top for the given user.}
    \label{fig:score_vs_iter}
\end{figure}

We evaluated the behavior of the system, taking the score of the proxy model, GP-BPR, at different recommendation steps. Each iteration depends on the result of the previous one, including also the personalized feedback evaluating the recommended outfit. In our experiments, we always limit the number of iterative recommendations to 10 for simplicity. However, this is not a limit posed by the model, which can recommend garments indefinitely. In a real scenario, we could expect the loop to terminate whenever the user is satisfied with the recommendation.
We report in Fig.~\ref{fig:score_vs_iter} the results for our reinforcement learning agent. We also report the results obtained by a similar agent that has been trained without exploration, i.e. using $\epsilon=0$ during the whole training phase (see Eq.~\ref{eq:epsilon}).
In addition, we use two baselines: LSTM and random. The former is a simple LSTM model with a hidden state of dimension 2048 (same as the image feature), that is initialized with the top garment feature and receives as input at every timestep the feedback score, lifted to the hidden dimension with a linear layer. A final linear layer selects the bottom garment to be proposed. This model is not trained with reinforcement learning but it is trained end-to-end with vanilla SGD.
The random baseline instead acts as a lower bound and simply draws a random bottom at every iteration. All results are averaged across every test sample.

The reinforcement learning agent, on average obtains a compatibility score that increases during the first five iterations and eventually saturates to high values. This indicates that the model is indeed capable of inferring user preferences on the fly and proposing garments that are increasingly more suitable for the given bottom and user.
Using the model without exploration instead, the recommendations tend to have oscillating yet similar scores, meaning that the model is not able to discover personalized outfits based on previous recommendations. This is due to the fact that during training the agent learns to propose a small set of garments that are good on average, thus limiting the personalization of the recommendation set. However, on average, both the models are consistently better than the random and LSTM baselines, confirming that even the model without exploration is capable of capturing general preferences, although without being able to personalize the recommendations.
In particular, it is interesting to observe that the LSTM model does not manage to produce meaningful recommendations and obtains on average very similar results to random. We found that training an LSTM-based model for this task has convergence issues since it always predicts the same bottoms, regardless of the user and the top. This once again confirms the importance of the exploration term in reinforcement learning agents, which allows a model to learn to recommend diverse items and take into account both the user and the top.

\begin{figure}[t]
    \centering
    \includegraphics[width=0.49\textwidth]{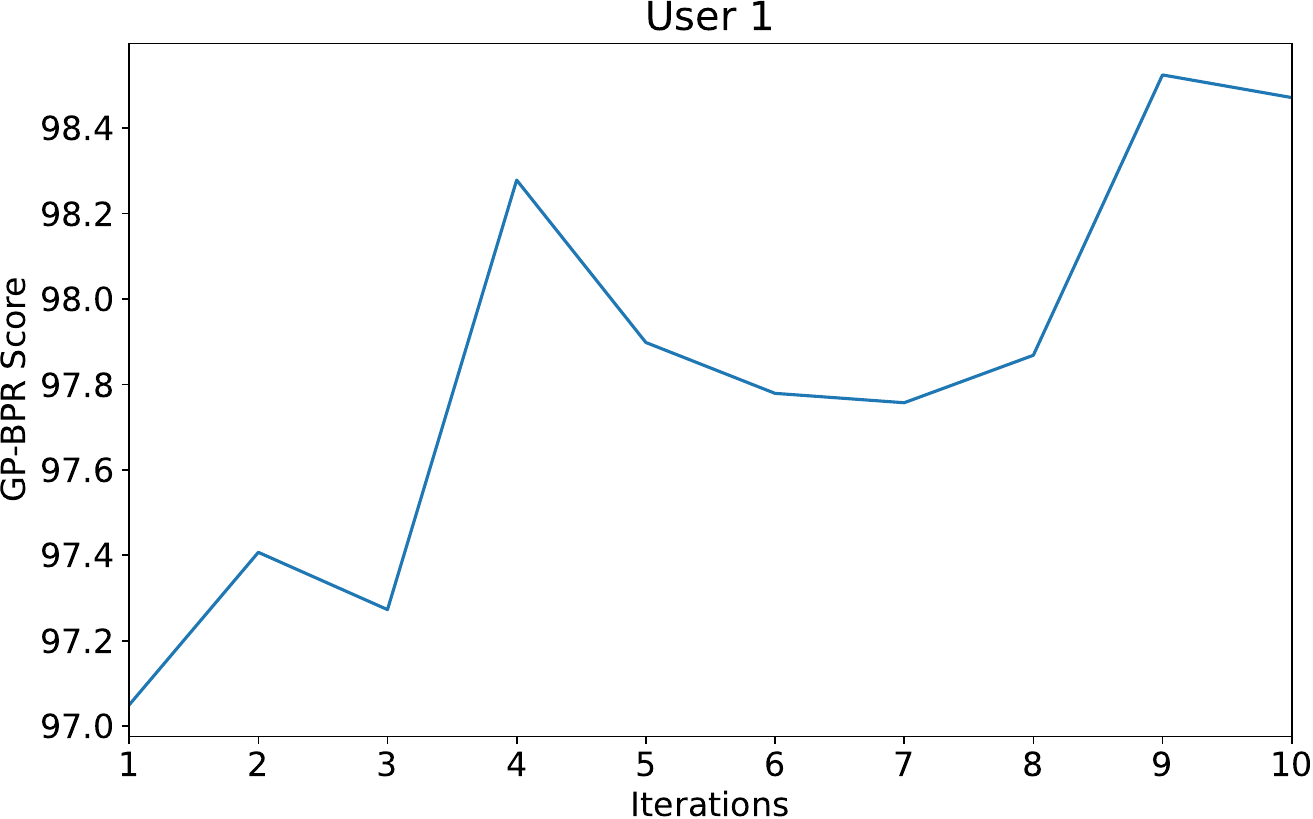}
    \includegraphics[width=0.49\textwidth]{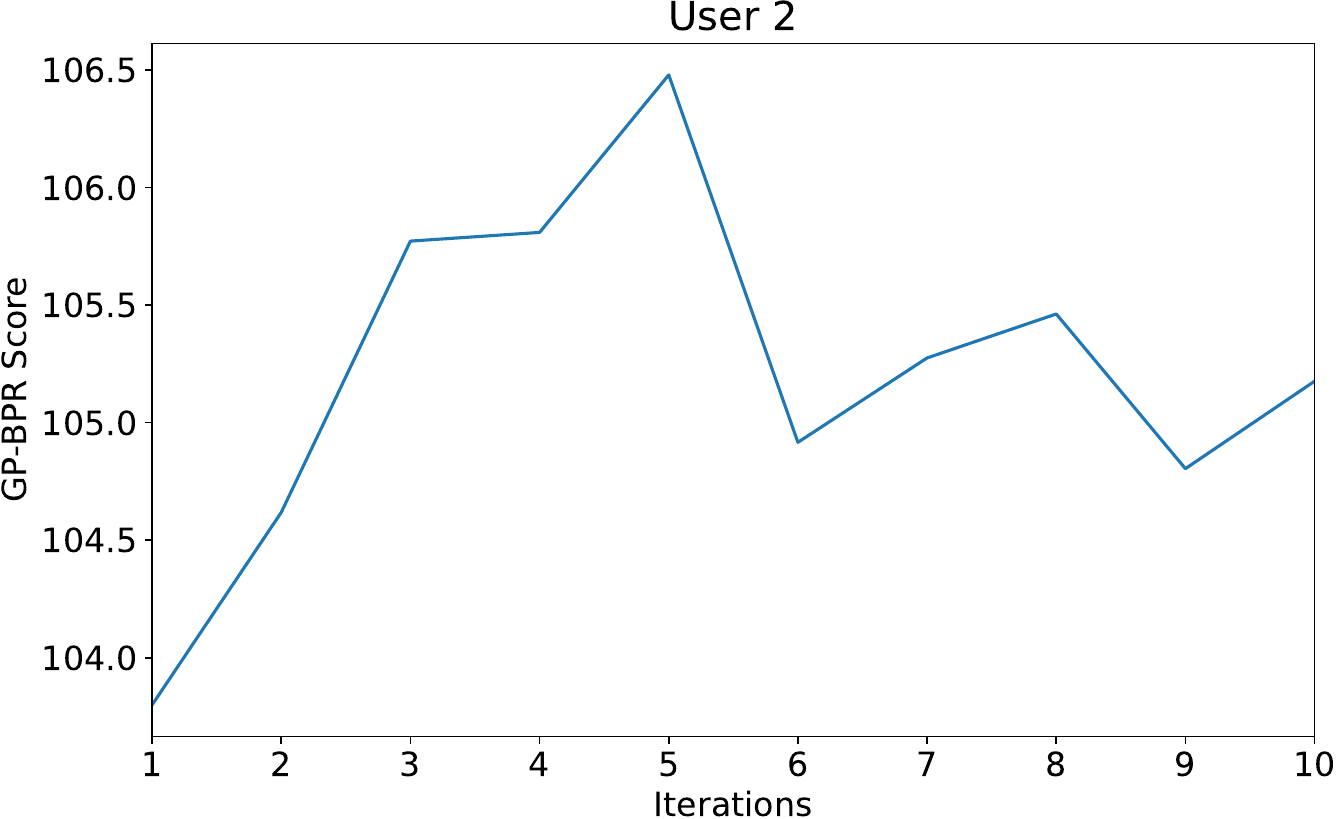}\\
    ~~~~~\\
    \includegraphics[width=0.49\textwidth]{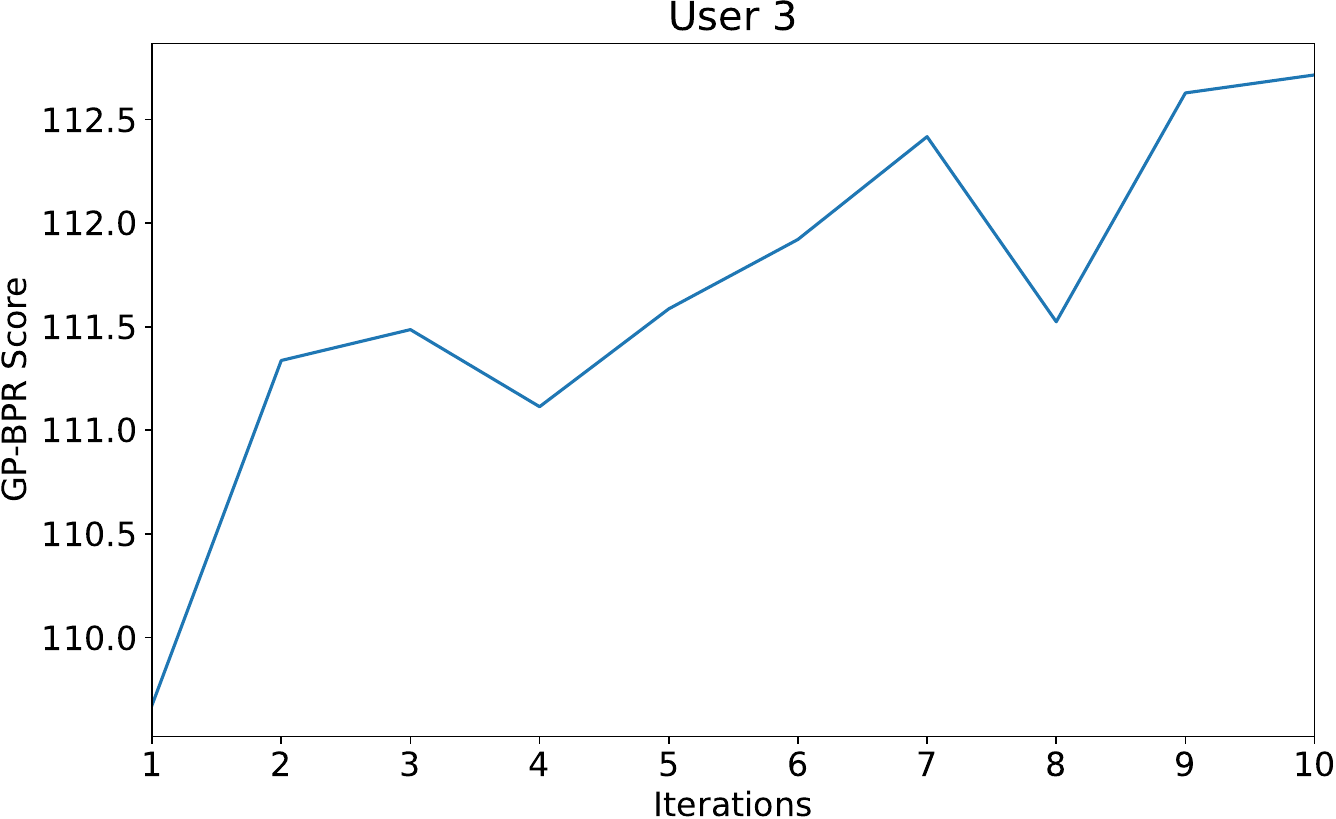}
    \includegraphics[width=0.49\textwidth]{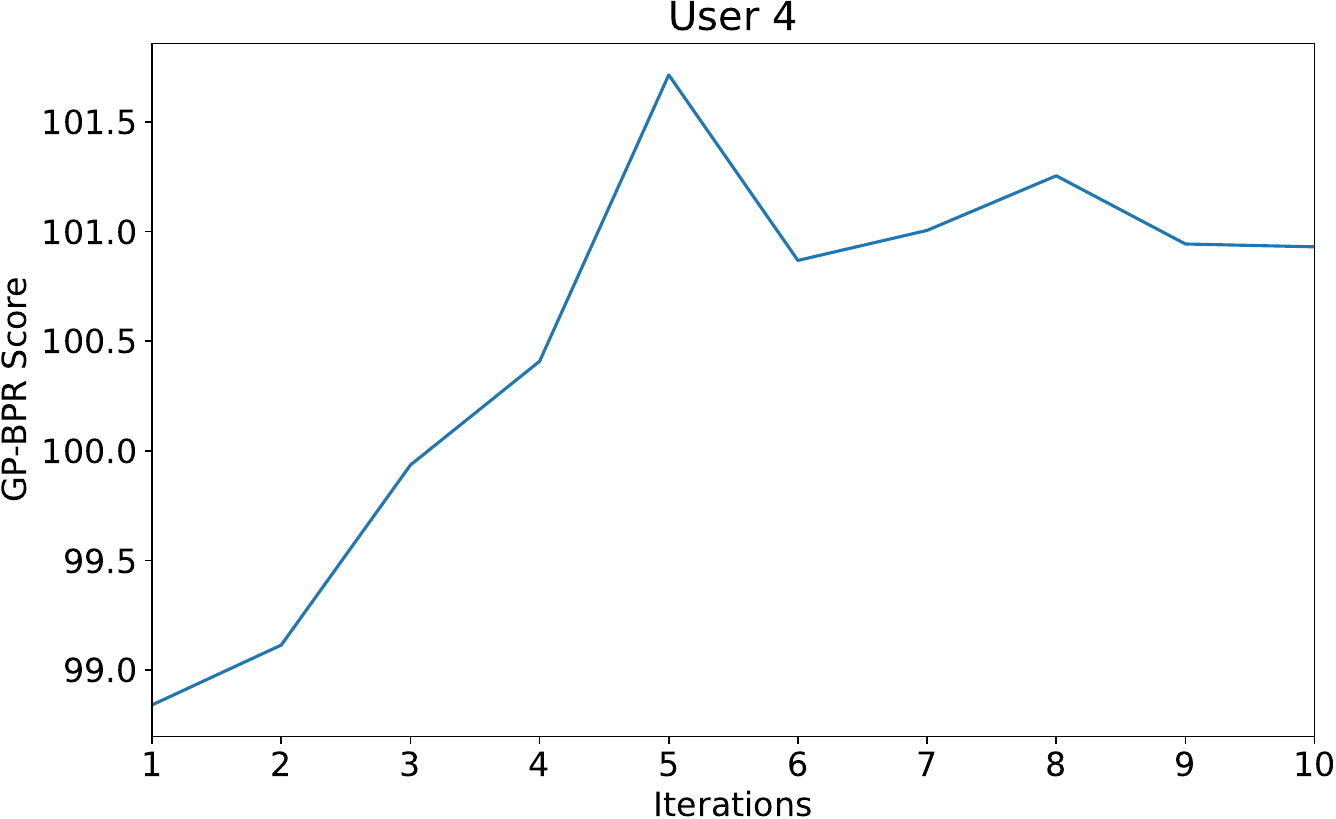}
    \caption{Trends of individual episodes. For each plot, we select a random user and fix a random top and we show the G-BPR scores of the recommended bottoms. The increasing trend is well observable in every plot.}
    \label{fig:score_iter_breakdown}
\end{figure}

To provide a better analysis of the behavior of the model, we analyze a few individual samples, randomly selected from the test set. Fig.~\ref{fig:score_iter_breakdown} shows four different samples relative to different users with different top garments. The increasing trend of the recommendation score is clearly observable in all samples. For \textit{User 2} and \textit{User 4}, the curves are similar to the average one of Fig.~\ref{fig:score_vs_iter}, while \textit{User 1} and \textit{User 2} reach the highest scoring recommendation in the final steps.
Interestingly, we can also observe that there is a bias in the scores provided by the proxy model. The bias is given by the combination of user and top, which leads to different ranges of scores from one sample to another. For example, some users always provided low scores in the training set and there are top garments that are on average well appreciated or disliked, regardless of the paired bottom. However, we can see that such bias is consistent within an episode and that different episodes exhibit similar increasing trends.

We then inspect the \textit{Higher-than-Negative} and \textit{Higher-than-Positive} metrics. Tab. \ref{tab:hn_hp} reports the results for the four methods, i.e. indicating the percentage of episode in which at least one recommendation is considered to be better than the positive ground truth (HP) or the negative reference (HN), averaged over the whole test set. Interestingly, our reinforcement learning agent is capable of suggesting at least one bottom that is more satisfying than the ground truth 65\% of the time. The baselines instead obtain lower results. In particular, the LSTM baseline performs worse than random, since it always proposes the same bottoms. We attribute the higher results of the random baseline to the fact that by chance the baseline could select garments that obtain a higher score than the ones selected by the LSTM. The reinforcement learning agent also consistently recommends bottoms that are better than the negative reference.

\begin{figure}[t]
    \centering
    \includegraphics[width=0.49\textwidth]{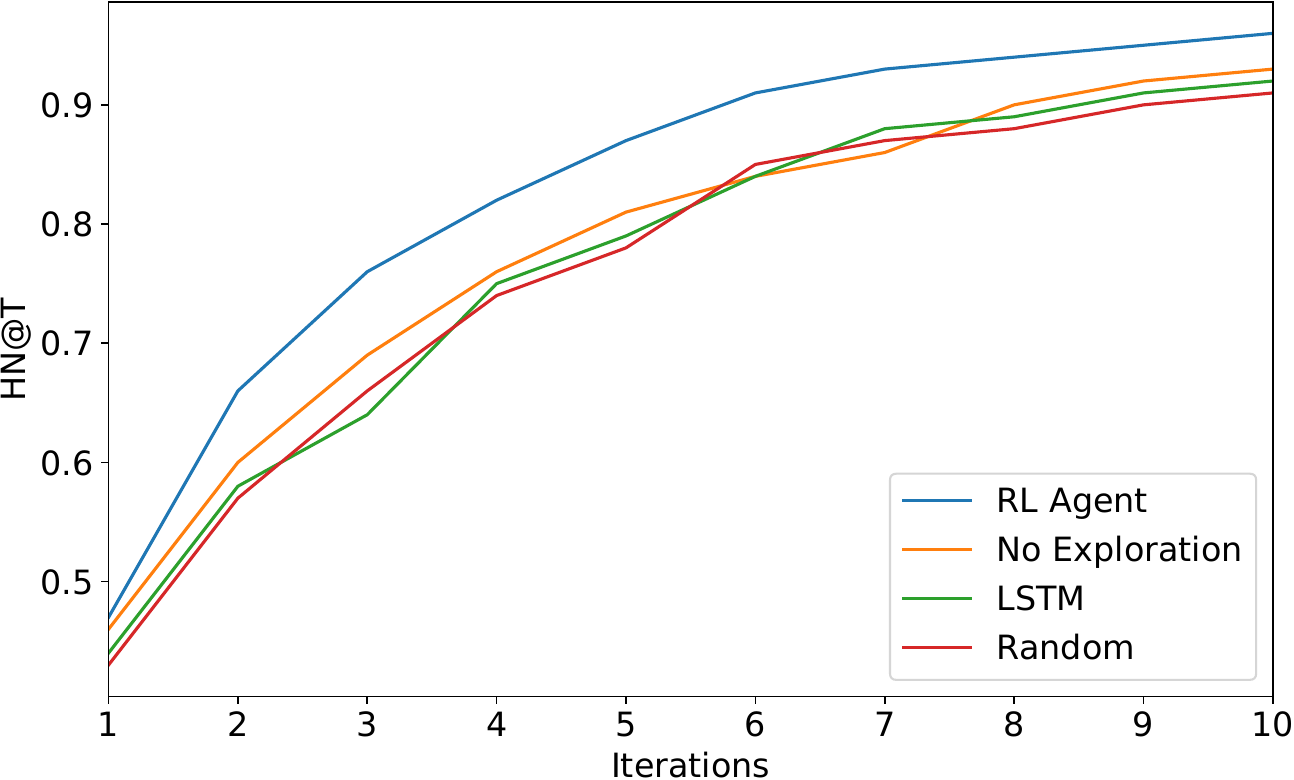}
    \includegraphics[width=0.49\textwidth]{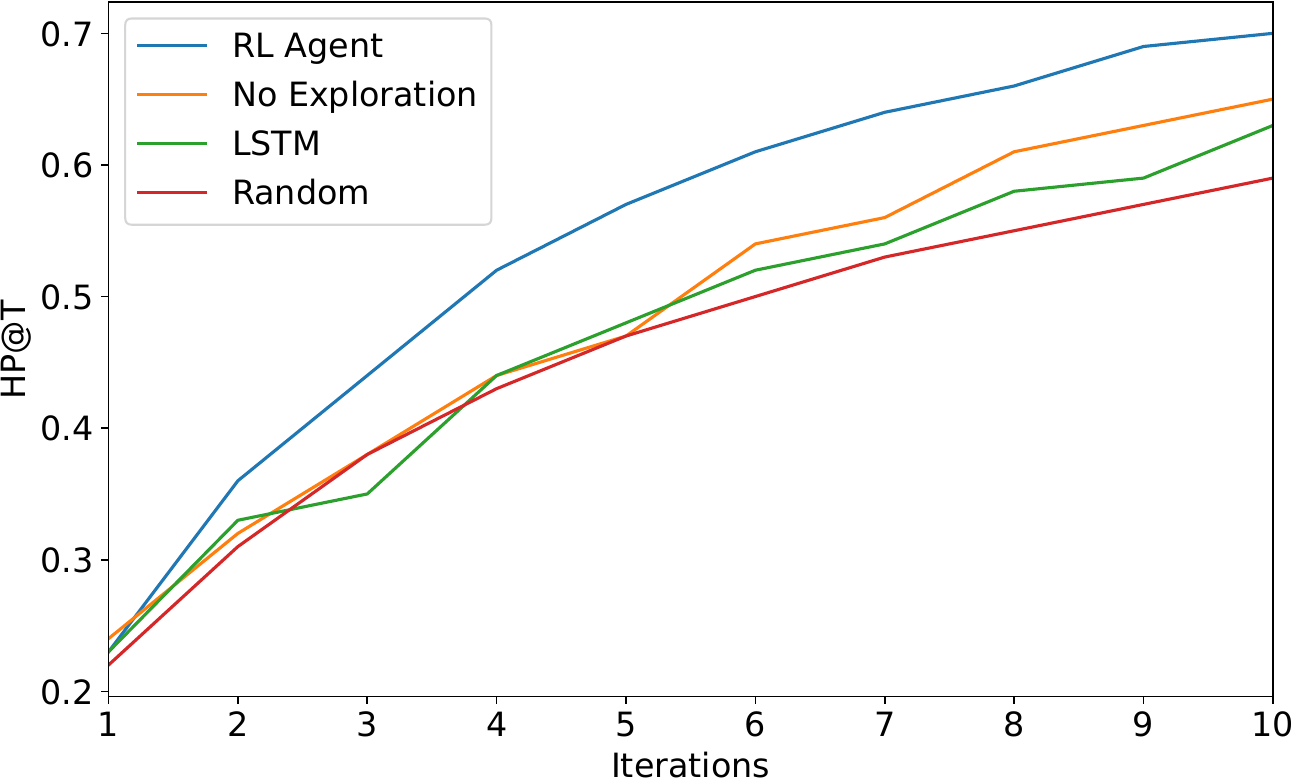}
    \caption{\review{HN@T and HP@T, for various T. We consider a recommended bottom to be a hit if there is at least one item better than the negative/positive ground truth items in the first T recommendations.}}
    \label{fig:hn_hp}
\end{figure}

\review{To see how the recommendations evolve through time, Fig. \ref{fig:hn_hp} shows the HN@T and HP@T metrics, for T=1,...,10, i.e. throughout the whole episode}. Results are averaged across all episodes in the test set. As expected both metrics increase monotonically for all methods, with the reinforcement learning agent obtaining the best scores for every iteration. This means that the agent is capable of modeling the preferences of the users earlier and more effectively than the others.

\begin{table}[t]
    \centering
    \begin{tabular}{c|ccccc}
    & RL Agent & No Exploration & LSTM & Random \\ \hline
         HN & 0.96 & 0.93 & 0.92 & 0.91 \\
         HP & 0.70 & 0.65 & 0.63 & 0.59
    \end{tabular}
    \caption{\textit{Higher-than-Negative} and \textit{Higher-than-Positive} metrics. These metrics indicate the percentage of episodes where the model is able to recommend at least one garment which is better than the negative reference (HN) and the positive ground truth (HP).}
    \label{tab:hn_hp}
\end{table}

Finally, we report statistics about the recommendations in Tab. \ref{tab:num_rec}. In particular, we report the absolute number of bottom garments that are recommended when evaluating the model on the whole test set (10 recommendations for each sample). Interestingly, the reinforcement learning agent recommends a large number of bottoms (658 out of a total of 1500), almost 12 times more than its counterpart without exploration (56 out of 1500). This once again confirms the importance of exploration during training.
For the LSTM model, the variability is even worse. In fact, for every episode, the model always suggests the same 10 bottoms. On the other hand, the random baseline achieves the highest variability, which however is not informative since garments are drawn casually among the 1500 candidates.

The table also reports the average number of recommendations per episode that are better than the negative bottom in the dataset. Interestingly, more than half of the recommendations of the reinforcement learning agent are better than the negative sample.

\begin{table}[t]
    \centering
    \begin{tabular}{c|ccccc}
    & RL Agent & No Exploration & LSTM & Random \\ \hline
         \# Proposed bottoms & 658 & 56 & 10 & 1500 \\
         \# Score > Neg Bottom & 5.3 & 4.4 & 2.7 & 3.8
    \end{tabular}
    \caption{Statistics for the proposed method and baselines regarding the overall number of proposed garments when evaluating the whole test set (each episode lasts 10 iterations) and average number of recommended garments per episode, which is better than the negative reference bottom.}
    \label{tab:num_rec}
\end{table}

\subsection{Qualitative Results}
\begin{figure}[t]
    \centering
    \includegraphics[width=0.8\textwidth]{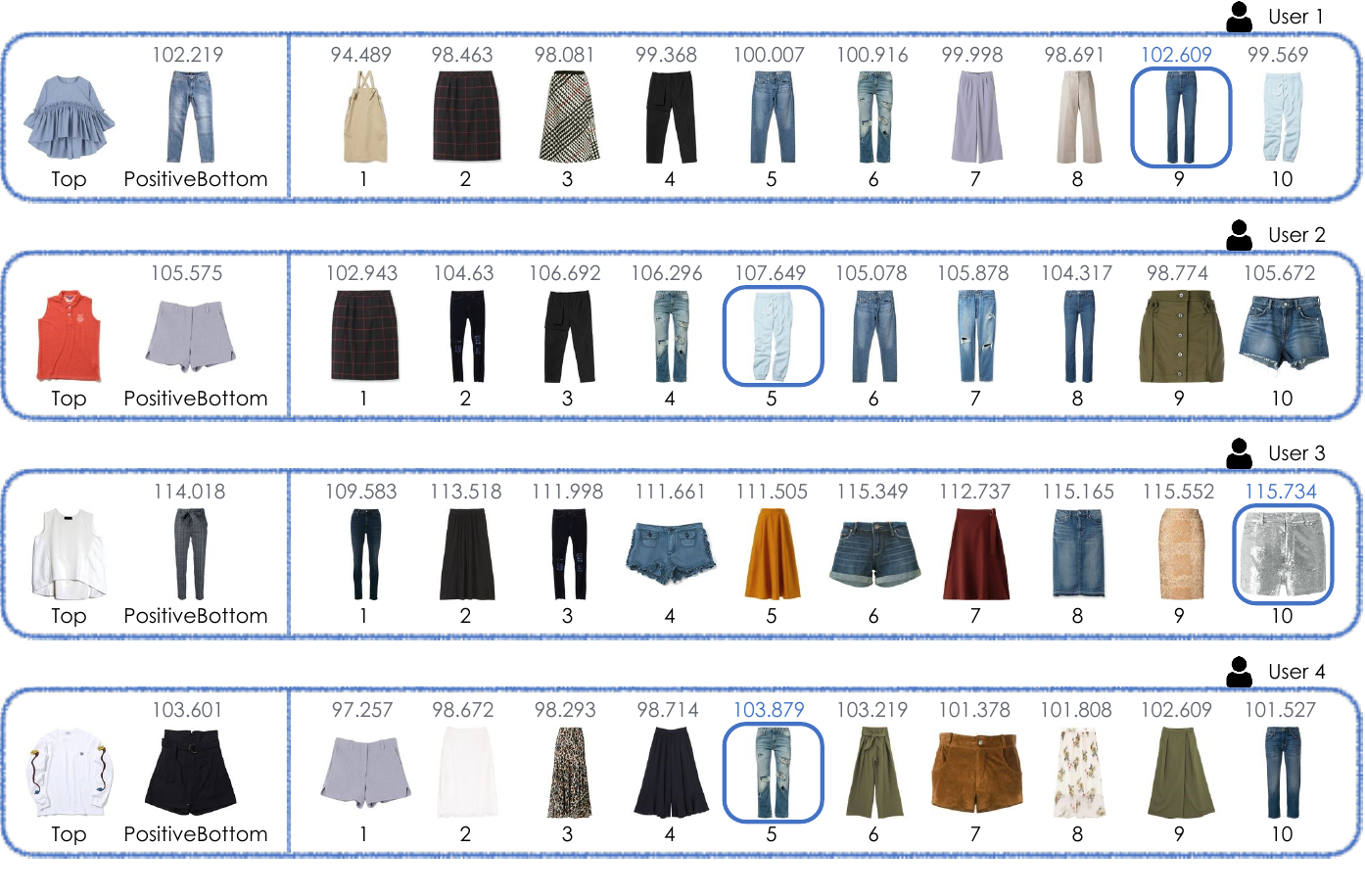}
    \caption{Qualitative results. We fix a given user and a top and suggest a sequence of bottom garments. For each recommended garment, we also show the unnormalized score provided by the proxy model GP-BPR. Highlighted in blue is the highest-scoring recommendation.}
    \label{fig:qualitative}
\end{figure}

We now provide a qualitative analysis of the results of the proposed approach.
In Fig. \ref{fig:qualitative} we show the results for four different episodes. For each episode, we fix a user and a top garment and we depict the positive bottom as well as its GP-BPR unnormalized compatibility score.
\review{The user is fixed by simply considering an annotated outfit in the test set for the given user (so that we have a ground truth reference) and we propose the top garment to our model. The user is then used to condition GP-BPR and get the estimated feedback at every iteration.}
At first, the recommendation agent suggests a bottom based on general compatibility, without any personalization. It can be seen that these samples obtain a compatibility that is usually lower than the one of the ground truth. As the model keeps suggesting and observing the feedback, the scores improve, until a bottom with a sufficiently high score is suggested. The highest scoring recommendation is highlighted with a blue border in Fig. \ref{fig:qualitative}. Interestingly, a garment with a higher score than the ground truth is always obtained in these examples. Furthermore, such a garment is not necessarily similar to the ground truth. This is due to the fact that different outfits can be suitable for the user and that the one annotated in the dataset might not be the overall preferred combination for that user.
It is also important to notice the high variability of garments proposed by the model, as would be expected given the results in Tab. \ref{tab:num_rec}.

\begin{figure}[t]
    \centering
    \includegraphics[width=0.8\textwidth]{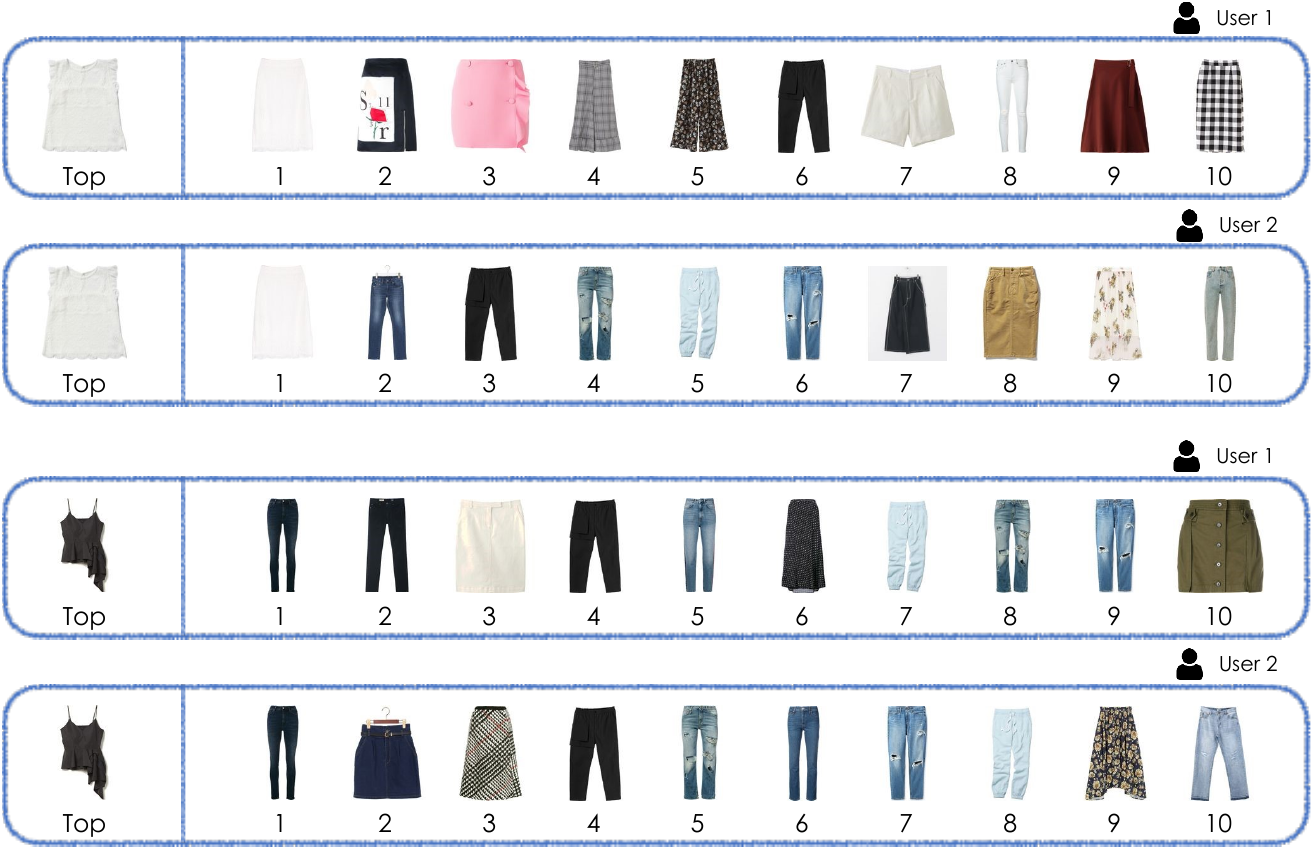}
    \caption{Recommendations for the same top, but for different users. The results are obtained with the reinforcement learning agent. It can be seen that the sequence of recommended bottoms changes based on the feedback of the user.}
    \label{fig:same_top}
\end{figure}

\begin{figure}[t]
    \centering
    \includegraphics[width=0.8\textwidth]{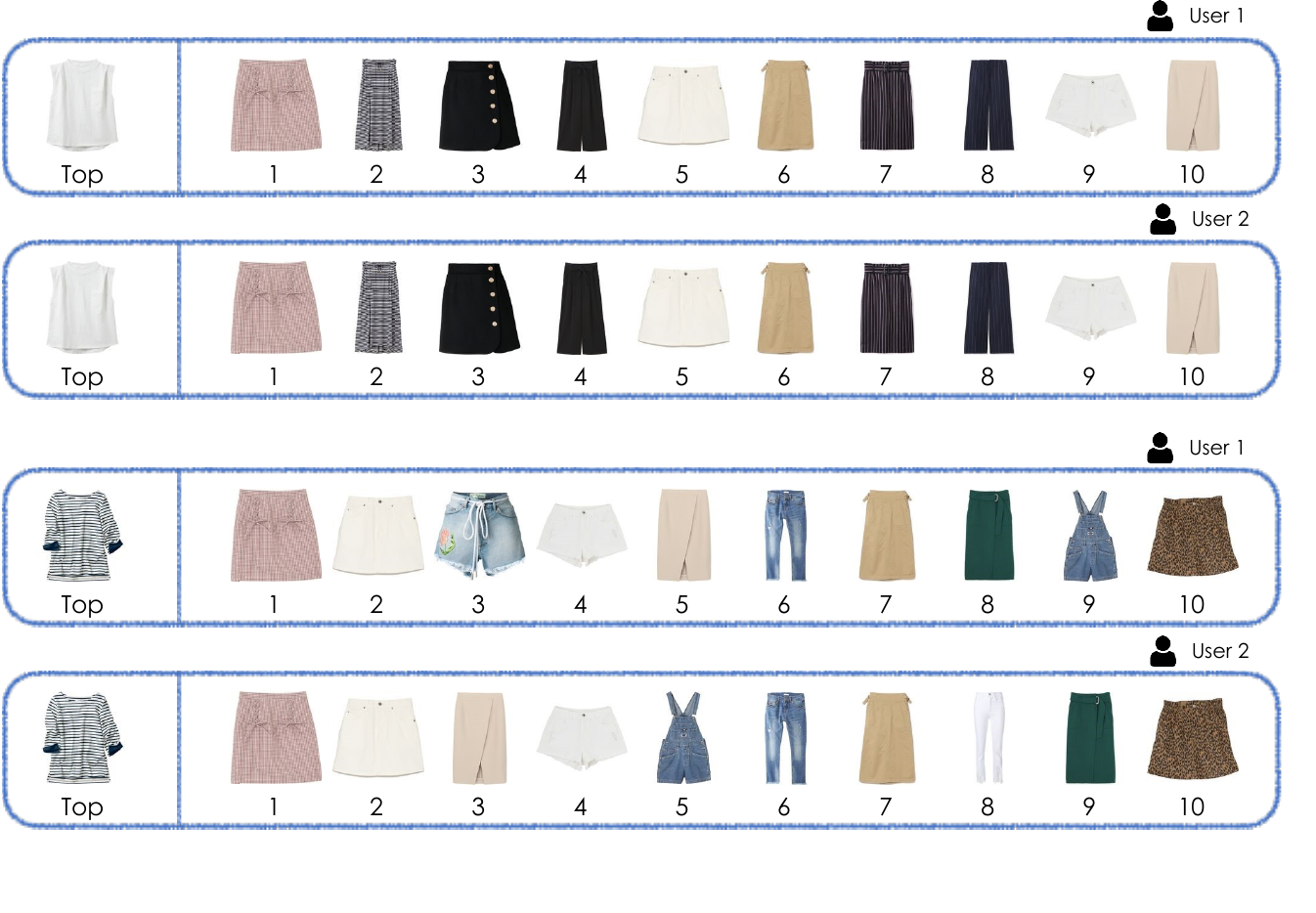}
    \caption{Recommendations for the same top, but for different users. The results are obtained with the reinforcement learning agent without exploration. In this case, the agent does not manage to diversity the recommendations with user preferences but only suggests bottoms that are suitable with the top. Garments are also often repeated for different tops.}
    \label{fig:same_top_no_exploration}
\end{figure}

To be sure that the model is actually personalizing the recommendations, we show in Fig. \ref{fig:same_top} and Fig. \ref{fig:same_top_no_exploration} the sequence of recommendations obtained by the reinforcement learning agent respectively with and without exploration. In this case, we fix the top garment and select two different users at random. It can be seen that the agent with exploration (Fig. \ref{fig:same_top}) starts with a recommendation that only depends on the bottom and then starts to diversify its outputs, based on the feedback of the user.
Differently, the agent trained without exploration (Fig. \ref{fig:same_top_no_exploration}) is less capable of personalizing the outputs. In the first example, all recommended items are the same for both users, whereas in the second one, we can note two interesting facts. First, the recommendation starts with the same outfit in all episodes, even if the top or the user change. Second, despite there being diversity from the prior episode and the two sequences are not the same, most of the recommended bottoms are the same but in a different order. This suggests that the agent without exploration is less inclined to change its recommendations based on user feedback and that there is a bias towards a small subset of bottoms.

% \todo{Qualitative results}
% \begin{itemize}
%     \item Example of sequence of outputs
%     \item Same top and different user yields to different bottom recommendations
% \end{itemize}

% ---------------------------------------------------------------------------------

% \section{Ablation Studies}
% \label{sec:ablation}
% \todo{We have some ablation studies for the RL model, e.g. without exploration}
% \todo{Define a list of ablations}

% ---------------------------------------------------------------------------------

\section{Conclusion and Future Works}
\label{sec:conclusion}
We have presented a system for personalized recommendations of compatible garments. In our framework, the model proposes a garment and waits for user feedback before providing the next recommendation. The aim is to maximize the satisfaction of the user, attempting to model user preferences without any prior knowledge about the user. We found reinforcement learning, and including an exploration term during training, to be fundamental to achieving our goal. Of particular importance when defining our training loop was to define a proxy model to approximate the feedback of the user, and therefore overcome the requirement of having actual humans in the loop.
As directions for future work, we intend to bring the paradigm of interactive garment recommendation closer to real use with human subjects, possibly fine-tuning the model directly in stores in an online learning fashion, to continuously improve by observing real feedback. We also intend to combine our approach with the recent developments in the field of vision and language, that is offering the possibility to the user to interact verbally with the model to perform explicit actions such as changing colors or defining a specific type of garment.

% ---------------------------------------------------------------------------------

% \begin{acks}

% \end{acks}

% ---------------------------------------------------------------------------------

\bibliographystyle{ACM-Reference-Format}
\bibliography{tomm}

\end{document}